# ABCNet: Attentive Bilateral Contextual Network for Efficient Semantic Segmentation of Fine-Resolution Remote Sensing Images

Rui Li[1] and Chenxi Duan[2, *]

1) School of Remote Sensing and Information Engineering, Wuhan University, 129 Luoyu Road, Wuhan, Hubei 430079, China.

2) The State Key Laboratory of Information Engineering in Surveying, Mapping and Remote Sensing, Wuhan University, 129 Luoyu Road, Wuhan, Hubei 430079, China.

E-mail addresses: lironui@whu.edu.cn (R. Li), chenxiduan@whu.edu.cn (C. Duan)

*Corresponding author.

*Abstract*—Semantic segmentation of remotely sensed images plays a crucial role in precision agriculture, environmental protection, and economic assessment. In recent years, substantial fine-resolution remote sensing images are available for semantic segmentation. However, due to the complicated information caused by the increased spatial resolution, state-of-the-art deep learning algorithms normally utilize complex network architectures for segmentation, which usually incurs high computational complexity. Specifically, the high-caliber performance of the convolutional neural network (CNN) heavily relies on fine-grained spatial details (fine resolution) and sufficient contextual information (large receptive fields), both of which trigger high computational costs. This crucially impedes their practicability and availability in real-world scenarios that require real-time processing. In this paper, we propose an Attentive Bilateral Contextual Network (ABCNet), a convolutional neural network (CNN) with double branches, with prominently lower computational consumptions compared to the cutting-edge algorithms, while maintaining a competitive accuracy. Code is available at https://github.com/lironui/ABCNet.

## 1. INTRODUCTION

Profit from the rapidly expanding Earth Observation technique, a large amount of remotely sensed images with fine spatial and spectral resolutions are now available for a wide range of application scenarios such as image classification (Lyons et al., 2018; Maggiori et al., 2016), object detection (Li et al., 2017; Xia et al., 2018), and semantic segmentation (Kemker et al., 2018; Zhang et al., 2019a). The revisiting property of orbital acquisitions brings the consecutive monitoring of land surface, ocean, and atmosphere into the possibility (Duan and Li, 2020). Fine-resolution remote sensing images normally contain substantial detailed spatial information for land cover and land use (Duan et al., 2020). Semantic segmentation, which assigns each pixel in images with a definite category, has become one of the most crucial levers for ground object interpretation. Specifically, semantic segmentation from remotely sensed imagery plays a pivotal role in various scenarios including precision agriculture (Griffiths et al., 2019; Picoli et al., 2018), environmental protection (Samie et al., 2020; Yin et al., 2018), and economic assessment (Zhang et al., 2020; Zhang et al., 2019a). Looking from a panoramic view, semantic segmentation is one of the high-level tasks that paves the way for complete scene understanding. Hence, semantic segmentation is at the forefront of a comprehensive effort towards automatic Earth monitoring by international agencies.

To identify the image content from various land cover and land use categories, tons of approaches explored the utilization of spectral and spectral-spatial features to interpret remote sensing images (Gong et al., 1992; Ma et al., 2017; Tucker, 1979; Zhong et al., 2014; Zhu et al., 2017). However, the finite ability to capture the contextual information contained in the images restricts the flexibility and adaptability of these methods (Li et al., 2020c; Tong et al., 2020), especially when the detailed and structural information surged by the increased spatial resolution. By contrast, bolstered by its powerful capabilities to capture nonlinear and hierarchical features automatically, deep Convolutional Neural Network (CNN) has posed a significant impact on the understanding of fine-resolution remote sensing images (Li et al., 2020a; Zheng et al., 2020).

For semantic segmentation, Fully Convolutional Network (FCN) (Long et al., 2015) is the first proven and effective end-to-end CNN structure. Restricted by the oversimple design of the decoder, the results of FCN, although very encouraging, appear coarse. Subsequently, the more elaborate encoder-decoder structure (Badrinarayanan et al., 2017; Ronneberger et al., 2015) is proposed which comprises two symmetric paths: a contracting path for extracting features and an expanding path for exact positioning to accomplish more accurate results. To guarantee the accuracy of segmentation, global contextual information and multiscale semantic features are supposed to be thoroughly utilized for semantic categories with varying sizes in images. By the spatial pyramid pooling module, the pyramid scene parsing network (PSPNet) (Zhao et al., 2017) aggregates contextual information among different regions. The dual attention network (DANet) (Fu et al., 2019) applies the dot-product attention mechanism to extract abundant contextual relationships. Subject to the enormous memory and computational consumptions, DANet simply attaches the dot-product attention mechanism at the lowest layer and merely captures the long-range dependencies from the smallest feature maps. DeeplabV3 (Chen et al., 2017) adopts atrous convolution to mining multiscale features, while a simple yet valid decoder module is added in DeepLabV3+ (Chen et al., 2018a) to further refine the segmentation results.

The extraction of global contextual information and the exploitation of large-scale feature maps are computationally expensive (Duan and Li, 2020; Li et al., 2020b). Therefore, a series of lightweight networks (Hu et al., 2020; Oršić and Šegvić, 2021; Romera et al., 2017; Yu et al., 2018; Zhuang et al., 2019) are developed to accelerate the computational speed while keeps the equilibrium between accuracy and efficiency. For example, the asymmetric convolution which is used in ERFNet (Romera et al., 2017) factorizes the standard $3 \times 3$ convolutions into a $1 \times 3$ convolution and a $3 \times 1$ convolution, saving about 33% computational consumptions. By exploiting spatial correlations and cross-channel correlations respectively, BiseNet (Yu et al., 2018) utilizes the depth-wise separable convolution (Chollet, 2017) which further lowers the consumption of the standard convolution. Multi-scale encoder-decoder branch pairs with skip connections are studied in ShelfNet (Zhuang et al., 2019) where a shared-weight strategy is harnessed in the residual block to reduces the parameter without sacrificing accuracy. For implementing the non-local context aggregation, FANet (Hu et al., 2020) employs the fast attention module in efficient semantic segmentation. SwiftNet (Oršić and Šegvić, 2021) explores the effectiveness of pyramidal fusion in compact architectures.

Due to limited capacity in extracting the global context information, there is a huge gap in accuracy between the lightweight networks and the state-of-the-art models, which is especially true when it comes to the fine-resolution remotely sensed images. As a powerful approach that can capture long-range dependencies, the dot-product attention mechanism (Vaswani et al., 2017) is a plausibly ideal solution to remedy this limitation. Whereas, the memory and computational consumptions of the dot-product attention mechanism increase quadratically with the spatio-temporal size of the input, which runs counter to the original intention of lightweight networks. Encouragingly, our previous work about linear attention (Li et al., 2020a) which reduces the complexity of the dot-product attention mechanism from $O(N^2)$ to $O(N)$ alleviates this plight.

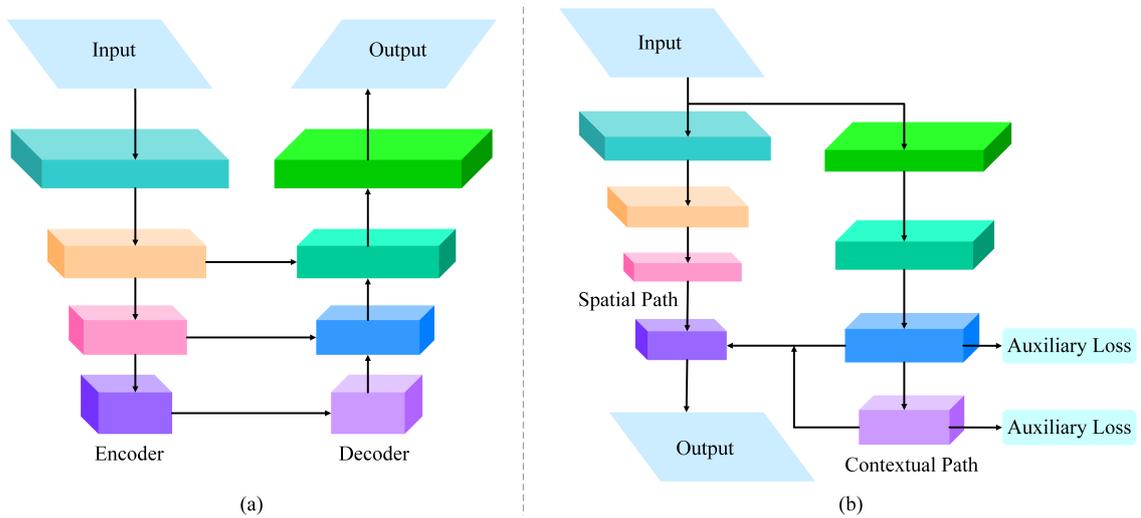

Fig.1 Illustration of (a) the encoder-decoder structure and (b) the bilateral architecture.

In this paper, we aim to further improve the segmentation accuracy while simultaneously ensuring the efficiency of semantic segmentation. We approach this challenging problem by modeling the global contextual information using the linear attention mechanism. To be specific, we proposed an Attentive Bilateral Contextual Network (ABCNet) to address the efficient semantic segmentation of fine-resolution remote sensing images. Following the design philosophy of BiSeNet (Yu et al., 2018), there are two branches in the proposed ABCNet: a spatial path to retain affluent spatial details and a contextual path to capture global contextual information.

Compared with the encoder-decoder structure (Fig. 1(a)), the bilateral architecture (Fig. 1(b)) can maintain more spatial information without retarding the speed of the model (Yu et al., 2018). Concretely, the spatial path merely stacks three convolution layers to generate the 1/8 feature maps, while the contextual path includes two attention enhancement modules (AEM) to refine the features and capture contextual information. As features generated by two paths are disparate in the level of feature representation, we further design a feature aggregation module (FAM) to fuse these features. Our main contributions are summarized as follows:

1) We propose a novel approach for efficient semantic segmentation of fine-resolution remote sensing images. Specifically, we propose an Attentive Bilateral Contextual Network (ABCNet) with a spatial path and a contextual path.

2) We design two specific modules, attention enhancement modules (AEM) for exploring long-range contextual information and feature aggregation module (FAM) for fusing features obtained by two paths.

3) We achieve competitive results on the ISPRS Vaihingen dataset and ISPRS Potsdam dataset. More specifically, we obtain the results of 91.095% overall accuracy on the Potsdam test dataset with a speed of 72.13 FPS even on a mid-range graphics card (1660Ti).

## 2. Related Work

### 1) Context information extraction

As the performance of semantic segmentation heavily hinges on the abundant context information, a great many endeavors are poured into tackling this issue. The dilated or atrous convolution (Chen et al., 2014; Yu and Koltun, 2015) has been demonstrated to be an effective technology for enlarging receptive fields without shrinking spatial resolution. Also, the encoder-decoder (Ronneberger et al., 2015) architecture which merges high-level and low-level features using skip connections is another valid way for extracting spatial context. Based on the encoder-decoder framework or dilation backbone, several subsequent studies focus on exploring the usage of spatial pyramid pooling (SPP) (He et al., 2015). For example, the pyramid pooling module (PPM) in PSPNet is composed of convolutions with kernels of four different sizes (Zhao et al., 2017), while DeepLab v2 (Chen et al., 2018a) equips with the atrous spatial pyramid pooling (ASPP) module which groups parallel atrous convolution layers with varying dilation rates. However, there are still certain current limitations in SPP. The SPP with standard convolution will face a dilemma when expanding the receptive field by a large kernel size. The above operations are normally accompanied by a huge number of parameters. The SPP with small kernels (e.g. ASPP), on the other hand, lacks enough connection between adjacent features; and the gridding problem (Wang et al., 2018a), which occurs when the field is enlarged by a dilated convolutional layer. By contrast, the powerful ability to model long-range dependencies enable the dot-product attention mechanism to extract context information in the global scale.

## 2)  Dot-Product Attention Mechanism

Let $H$, $W$, and $C$ denote the height, weight, and channels of the input, respectively. The input feature is defined as $\boldsymbol{X} = [\boldsymbol{x_1}, \cdots, \boldsymbol{x_N}] \in \mathbb{R}^{N \times C}$, where $N = H \times W$. Firstly, the dot-product attention mechanism utilizes three projected matrices $\boldsymbol{W_q} \in \mathbb{R}^{D_x \times D_k}$, $\boldsymbol{W_k} \in \mathbb{R}^{D_x \times D_k}$, and $\boldsymbol{W_v} \in \mathbb{R}^{D_x \times D_v}$ to generate the corresponding query matrix $\boldsymbol{Q}$, the key matrix $\boldsymbol{K}$, and the value matrix $\boldsymbol{V}$:

$$\begin{cases} \boldsymbol{Q} = \boldsymbol{XW}_q \in \mathbb{R}^{N \times D_k}; \\ \boldsymbol{K} = \boldsymbol{XW}_k \in \mathbb{R}^{N \times D_k}; \\ \boldsymbol{V} = \boldsymbol{XW}_v \in \mathbb{R}^{N \times D_v}. \end{cases} \qquad (1)$$

Please note that the dimensions of the $\boldsymbol{Q}$ and $\boldsymbol{K}$ are supposed to be identical and all the vectors in this section are column vectors by default. Accordingly, a normalization function $\rho$ is employed to measure the similarity between the $i$-th *query* feature $\boldsymbol{q}_i^T \in \mathbb{R}^{D_k}$ and the $j$-th *key* feature $\boldsymbol{k}_j \in \mathbb{R}^{D_k}$ as $\rho(\boldsymbol{q}_i^T \boldsymbol{k}_j) \in \mathbb{R}^1$. As the *query* feature and *key* feature are generated via different layers, the similarities between $\rho(\boldsymbol{q}_i^T \boldsymbol{k}_j)$ and $\rho(\boldsymbol{q}_j^T \boldsymbol{k}_i)$ are not symmetric. By calculating similarities between all pairs of pixels in the input feature maps and taking the similarities as weights, the dot-product attention mechanism generates the value at position $i$ by aggregating the *value* features from all positions using weighted summation:

$$D(\boldsymbol{Q}, \boldsymbol{K}, \boldsymbol{V}) = \rho(\boldsymbol{QK}^T)\boldsymbol{V}. \qquad (2)$$

Normally, the softmax is the frequently-used normalization function:

$$\rho(\boldsymbol{Q}^T\boldsymbol{K}) = softmax_{row}(\boldsymbol{QK}^T), \qquad (3)$$

where $softmax_{row}$ indicates that the softmax is exploited along each row of the matrix $\boldsymbol{QK}^T$.

By modeling the similarities between each pair of positions of the input, the global dependencies in the features can be thoroughly extracted by the $\rho(\boldsymbol{QK}^T)$. The dot-product attention mechanism is firstly designed for machine translation (Vaswani et al., 2017), while the non-local module (Wang et al., 2018b) introduces and modifies it for computer vision (Fig. 2). Based on the dot-product attention mechanism as well as its variants, a constellation of attention-based networks has been proposed to tackle the semantic segmentation task. Inspired by the non-local module (Wang et al., 2018b), the Double Attention Networks ($A^2$-Net) (Chen et al., 2018b), Dual Attention Network (DANet) (Fu et al., 2019), Point-wise Spatial Attention Network (PSANet) (Zhao et al., 2018), Object Context Network (OCNet) (Yuan and Wang, 2018), and Co-occurrent Feature Network (CFNet) (Zhang et al., 2019b) are proposed successively for scene segmentation by exploring the long-range dependency.

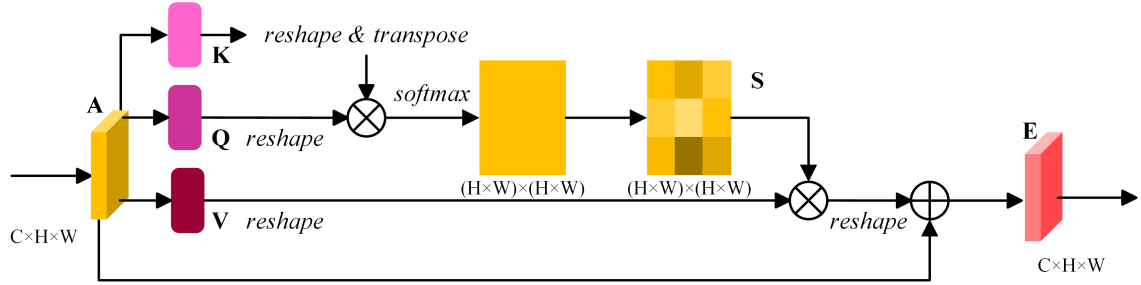

Fig.2 The diagram of the dot-product attention modified for computer vision.

Even though the introduction of attention significantly boosts the performance on segmentation, the huge resource-demanding of dot-product critically hinders its application on large inputs. To be specific, for $Q \in \mathbb{R}^{N \times D_k}$ and $K^T \in \mathbb{R}^{D_k \times N}$, the product between $Q$ and $K^T$ belongs to $\mathbb{R}^{N \times N}$, leading to the $O(N^2)$ memory and computation complexity. Consequently, it is requisite to lower the high demand for computational resources of the dot-product attention mechanism.

### 3) Generalization and simplification of the dot-product attention mechanism

If the normalization function is set as softmax, the *i*-th row of the result matrix generated by the dot-product attention mechanism can be written as:

$$D(Q, K, V)_i = \frac{\sum_{j=1}^{N} e^{q_i^T k_j} v_j}{\sum_{j=1}^{N} e^{q_i^T k_j}}. \tag{4}$$

Equation (4) can be rewritten and generalized to any normalization function as:

$$D(Q, K, V)_i = \frac{\sum_{j=1}^{N} \text{sim}(q_i, k_j) v_j}{\sum_{j=1}^{N} \text{sim}(q_i, k_j)}, \tag{5}$$
$$\text{sim}(q_i, k_j) \geq 0.$$

$\text{sim}(q_i, k_j)$ can be expanded as $\phi(q_i)^T \varphi(k_j)$ that measures the similarity between the $q_i$ and $k_j$, whereupon equation (4) can be rewritten as equation (6) and be simplified as equation (7):

$$D(\boldsymbol{Q}, \boldsymbol{K}, \boldsymbol{V})_i = \frac{\sum_{j=1}^{N} \phi(\boldsymbol{q}_i)^T \varphi(\boldsymbol{k}_j) \boldsymbol{v}_j}{\sum_{j=1}^{N} \phi(\boldsymbol{q}_i)^T \varphi(\boldsymbol{k}_j)}, \tag{6}$$

$$D(\boldsymbol{Q}, \boldsymbol{K}, \boldsymbol{V})_i = \frac{\phi(\boldsymbol{q}_i)^T \sum_{j=1}^{N} \varphi(\boldsymbol{k}_j) \boldsymbol{v}_j^T}{\phi(\boldsymbol{q}_i)^T \sum_{j=1}^{N} \varphi(\boldsymbol{k}_j)}. \tag{7}$$

Particularly, if $\phi(\cdot) = \varphi(\cdot) = e^{(\cdot)}$, equation (5) is equivalent to equation (4). The vectorized form of equation (7) is:

$$D(\boldsymbol{Q}, \boldsymbol{K}, \boldsymbol{V}) = \frac{\phi(\boldsymbol{Q}) \varphi(\boldsymbol{K})^T \boldsymbol{V}}{\phi(\boldsymbol{Q}) \sum_j \varphi(\boldsymbol{K})_{i,j}^T}. \tag{8}$$

As the softmax function is substituted for $\text{sim}(\boldsymbol{q}_i, \boldsymbol{k}_j) = \phi(\boldsymbol{q}_i)^T \varphi(\boldsymbol{k}_j)$, the order of the commutative operation can be altered, thereby avoiding multiplication between the reshaped *key* matrix $\boldsymbol{K}$ and *query* matrix $\boldsymbol{Q}$. In concrete terms, the product between $\varphi(\boldsymbol{K})^T$ and $\boldsymbol{V}$ can be computed first and then multiply the result and $\boldsymbol{Q}$, leading only $O(dN)$ time complexity and $O(dN)$ space complexity. The suitable $\phi(\cdot)$ and $\varphi(\cdot)$ enable the above scheme to achieve the competitive performance with finite complexity (Katharopoulos et al., 2020; Li et al., 2020b).

## 4) Linear Attention Mechanism

In our previous work (Li et al., 2020a) we proposed a linear attention mechanism from another perspective that replaces the softmax function with the first-order approximation of Taylor expansion, which is shown as equation (9):

$$e^{\boldsymbol{q}_i^T \boldsymbol{k}_j} \approx 1 + \boldsymbol{q}_i^T \boldsymbol{k}_j. \tag{9}$$

To guarantee the above approximation to be nonnegative, $\boldsymbol{q}_i$ and $\boldsymbol{k}_j$ are normalized by $l_2$ norm, thereby ensuring $\boldsymbol{q}_i^T \boldsymbol{k}_j \geq -1$:

$$sim(\boldsymbol{q}_i, \boldsymbol{k}_j) = 1 + \left(\frac{\boldsymbol{q}_i}{\|\boldsymbol{q}_i\|_2}\right)^T \left(\frac{\boldsymbol{k}_j}{\|\boldsymbol{k}_j\|_2}\right). \tag{10}$$

Thus, equation (5) can be rewritten as equation (11) and simplified as equation (12):

$$D(Q,K,V)_i = \frac{\sum_{j=1}^N \left(1 + \left(\frac{q_i}{\|q_i\|_2}\right)^T \left(\frac{k_j}{\|k_j\|_2}\right)\right) v_j}{\sum_{j=1}^N \left(1 + \left(\frac{q_i}{\|q_i\|_2}\right)^T \left(\frac{k_j}{\|k_j\|_2}\right)\right)}, \tag{11}$$

$$D(Q,K,V)_i = \frac{\sum_{j=1}^N v_j + \left(\frac{q_i}{\|q_i\|_2}\right)^T \sum_{j=1}^N \left(\frac{k_j}{\|k_j\|_2}\right) v_j^T}{N + \left(\frac{q_i}{\|q_i\|_2}\right)^T \sum_{j=1}^N \left(\frac{k_j}{\|k_j\|_2}\right)}. \tag{12}$$

The equation (12) can be turned into a vectorized form:

$$D(Q,K,V) = \frac{\sum_j V_{i,j} + \left(\frac{Q}{\|Q\|_2}\right)\left(\left(\frac{K}{\|K\|_2}\right)^T V\right)}{N + \left(\frac{Q}{\|Q\|_2}\right)\sum_j \left(\frac{K}{\|K\|_2}\right)^T_{i,j}}. \tag{13}$$

Since $\sum_{j=1}^N \left(\frac{k_j}{\|k_j\|_2}\right) v_j^T$ and $\sum_{j=1}^N \left(\frac{k_j}{\|k_j\|_2}\right)$ can be calculated and reused for each query, time and memory complexity of the attention based on equation (13) is $O(dN)$.

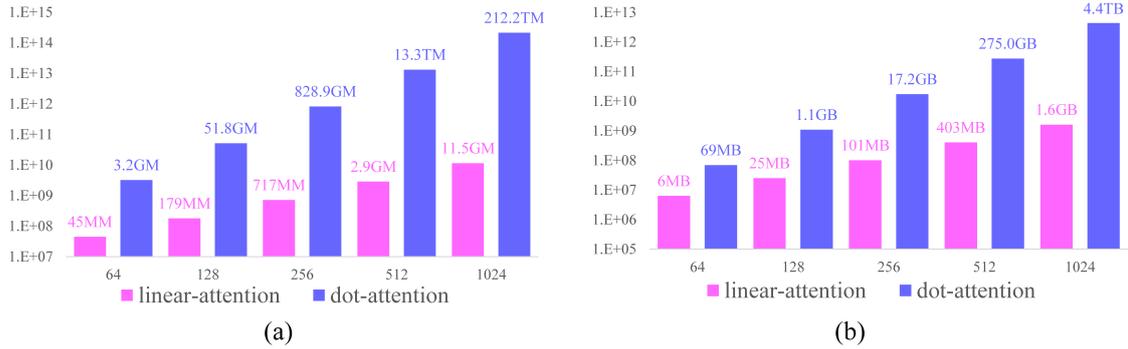

(a)                                           (b)

Fig.3 The (a) computation requirement and (b) memory requirement between the linear attention mechanism and dot-product attention mechanism under different input sizes. The calculation assumes $C = D_v = 2D_k = 64$. Please notice that the figure is on the log scale.

The validity and efficiency of the proposed attention have been testified through extensive ablation experiments and analysis (Li et al., 2020a).

## 5) Efficient semantic segmentation

For many applications, efficiency is critical, which is especially true for real-time ($\geqslant$30FPS)

scenarios such as autonomous driving. Therefore, recent researches have made great efforts to accelerate models for efficient semantic segmentation, which employs lightweight models or downsampling the input size. The utilization of lightweight convolutions (e.g., the asymmetric convolution and the depth-wise separable convolution) is a common strategy for designing lightweight networks (Romera et al., 2017; Yu et al., 2018). The downsampling of the input size is a trivial solution to speed up semantic segmentation which reduces the resolution of the input images, thereby leading to the loss of image details. To extract spatial details at original resolution, many methods further add a shallow branch, forming the two-path architecture (Yu et al., 2020; Yu et al., 2018).

## 3. Attentive Bilateral Contextual Network

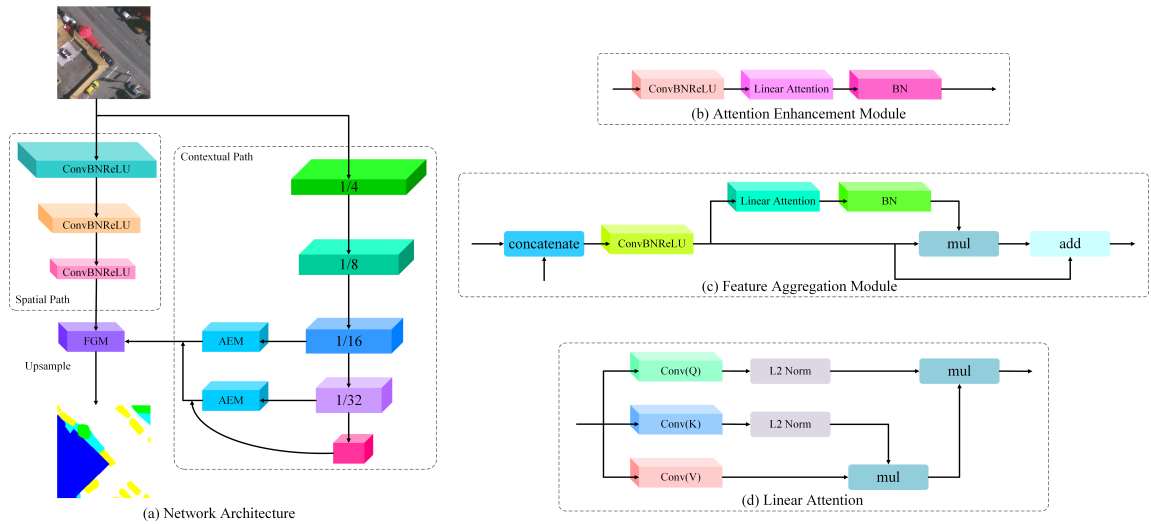

Fig.4 An overview of the Attentive Bilateral Contextual Network. (a) Network Architecture. (b) The Attention Enhancement Module (AEM). (c) The Feature Aggregation Module (FAM). (d) The Linear Attention Mechanism.

The proposed Attentive Bilateral Contextual Network (ABCNet), as well as the components,

are demonstrated in Fig. 4.

## 1) Spatial path

Although both of them are crucial for the high accuracy of segmentation, it is actually impossible to reconcile the affluent spatial details with the large receptive field simultaneously. Especially, in the term of efficient semantic segmentation, the mainstream solutions focus on down-sampling the input image or speeding up the network by channel pruning. The former loses the majority of spatial details, which the latter damages spatial details. By contrast, in the proposed ABCNet, we adopt the bilateral architecture (Yu et al., 2018) which is equipped with a spatial path to capture spatial details and generate low-level feature maps. Therefore, the rich channel capacity is essential for this path to encode sufficient spatial detailed information. Meanwhile, as the spatial path merely focuses on the low-level details, the shallow structure with a small stride for this branch is enough.

Specifically, the spatial path comprises three layers as shown in Fig. 4(a). Each layer contains a convolution with stride = 2, followed by batch normalization (Ioffe and Szegedy, 2015) and ReLU (Glorot et al., 2011). Therefore, the output feature maps of this path are 1/8 of the original image, which encodes abundant spatial details resulting from the large spatial size.

## 2) Contextual path

In parallel to the spatial path, the contextual path is designed to extract high-level global context information and provide sufficient receptive field. To enlarge the receptive field, several networks take advantage of the spatial pyramid pooling with a large kernel, leading to the huge computation demanding and memory consuming. With the consideration of the long-range context information and efficient computation simultaneously, we develop the contextual path with the linear attention mechanism (Li et al., 2020a).

Concretely, in the contextual path as shown in Fig. 4(a), we harness the lightweight backbone (i.e., ResNet 18) (He et al., 2016) to down-sample the feature map and encode the high-level semantic information. Thereafter, we deploy two attention enhancement modules (AEM) on the tails of the backbone to fully extract the global context information. The features obtained by the last two stages are fused and fed into the feature aggregation module (FAM).

## 3) Feature aggregation module

The feature representation of the spatial path and the contextual path is complementary but in different domains (i.e., the spatial path generates the low-level and detailed feature, while the contextual path obtains the high-level and semantic features). Thus, the simple fusion schemes such as summation and concatenation are not appropriate manners to fuse information. In contrast, we design a feature aggregation module (FAM) to merge both types of feature representation with consideration of accuracy and efficiency.

As shown in Fig. 4(c), with two domains of features, we first concatenate the output of spatial path and context path. Thereafter, a convolution layer with batch normalization (Ioffe and Szegedy, 2015) and ReLU (Glorot et al., 2011) attached to balance the scales of the features. Then, we capture the long-range dependencies of the generated features using the linear attention mechanism. The details of the design of FAM can be seen in Fig. 4(c).

**4) Loss function**

As can be seen from Fig. 1(b), besides the principal loss function to supervise the output of the whole network, we utilize two auxiliary loss functions at the context path to accelerate the convergence velocity. We select the cross-entropy loss as the principal loss:

$$loss_{pri}(p, y) = -y \log(p) - (1 - y) \log(1 - p), \tag{14}$$

where $p$ is the prediction generated by the network, while $y$ is the ground truth. The auxiliary loss functions are chosen as the focal loss:

$$loss_{aux1}(p, y) = loss_{aux2}(p, y) = -y(1 - p)^\gamma \log p - (1 - y)p^\gamma \log(1 - p), \tag{15}$$

where $\gamma$ is the focusing parameter, which controls the down-weighting of the easily classified examples and is set as 2 in our experiments. Hence, the overall loss of the network is:

$$loss(p, y) = loss_{pri} + loss_{aux1}(p, y) + loss_{aux2}(p, y). \tag{16}$$

## 4. EXPERIMENTAL RESULTS AND DISCUSSION

**1) Datasets**

The effectiveness of the proposed ABCNet is verified using the ISPRS Potsdam dataset, the ISPRS Vaihingen dataset.

**Potsdam**: There are 38 fine-resolution images of size $6000 \times 6000$ pixels with a ground sampling distance (GSD) of 5 cm in the Potsdam dataset. The dataset provides near-infrared, red, green, and blue channels as well as DSM and normalized DSM (NDSM). We utilize ID: 2_13, 2_14, 3_13, 3_14, 4_13, 4_14, 4_15, 5_13, 5_14, 5_15, 6_13, 6_14, 6_15, 7_13 for testing, ID: 2_10 for validation, and the remaining 22 images, except image named 7_10 with error annotations, for training. Please note that we only employ the red, green, and blue channels in our experiments.

**Vaihingen**: The Vaihingen dataset contains 33 images with an average size of 2494 × 2064 pixels and a GSD of 5 cm. The near-infrared, red, and green channels together with DSM are provided in the dataset. We utilize ID: 2, 4, 6, 8, 10, 12, 14, 16, 20, 22, 24, 27, 29, 31, 33, 35, 38 for testing, ID: 30 for validation, and the remaining 15 images for training. The DSM is not used in our experiments.

## 2) Evaluation Metrics

The performance of ABCNet is evaluated using the overall accuracy (OA), the mean Intersection over Union (mIoU), and the F1 score (F1). Based on the accumulated confusion matrix, the OA, mIoU, and F1 are computed as:

$$OA = \frac{\sum_{k=1}^{N} TP_k}{\sum_{k=1}^{N} TP_k + FP_k + TN_k + FN_k}, \tag{17}$$

$$mIoU = \frac{1}{N} \sum_{k=1}^{N} \frac{TP_k}{TP_k + FP_k + FN_k}, \tag{18}$$

$$F1 = 2 \times \frac{precision \times recall}{precision + recall}, \tag{19}$$

where $TP_k$, $FP_k$, $TN_k$, and $FN_k$ represent the true positive, false positive, true negative, and false negatives, respectively, for object indexed as class $k$. OA is computed for all categories including the background.

## 3) Experimental Setting

All of the training procedures are implemented with PyTorch on a single Tesla V100 with 32 batch size, and the optimizer is set as AdamW with a 0.0003 learning rate. For training, the raw images are cropped into 512 × 512 patches and augmented by rotating, resizing, horizontal axis flipping, vertical axis flipping, and adding random noise. The comparative methods include the contextual information aggregation methods designed initially for natural images, such as pyramid scene parsing network (PSPNet) (Zhao et al., 2017) and dual attention network (DANet) (Fu et al., 2019), the multi-scale feature aggregation models proposed for remote sensing images, like multi-stage attention ResU-Net (MAResU-Net) (Li et al., 2020a) and edge-aware neural network (EaNet) (Zheng et al., 2020), and also lightweight network developed for efficient semantic segmentation including depth-wise asymmetric bottleneck network (DABNet) (Li et al., 2019), efficient residual factorized convNet (ERFNet) (Romera et al., 2017), bilateral segmentation network V1 (BiSeNetV1) (Yu et al., 2018) and V2 (BiSeNetV2) (Yu et al., 2020), fast attention network (FANet) (Hu et al., 2020), ShelfNet (Zhuang et al., 2019), and SwiftNet (Oršić and Šegvić, 2021). The test time augmentation (TTA) in terms of rotating and flipping is applied for all comparative methods.

## 4) Ablation study

To verify the effectiveness of the components in the proposed ABCNet, we conduct extensive ablation experiments. atmosphere conditions, while the setting details and quantitative results are listed in Table 1.

*Baseline*: We utilize the ResNet-18 as the backbone of the contextual path and select the contextual path without the AEM (denoted as *CP* in Table I) as the baseline. The feature maps generated by CP are directly up-sampled to the shape as the original input image.

*Ablation for attention enhancement module*: For capturing the global context information, we

TABLE I

ABLATION STUDY OF EACH COMPONENT IN OUR PROPOSED ABCNET

| Dataset | Method | Mean F1 | OA (%) | mIoU (%) |
|---------|--------|---------|--------|----------|
| Vaihingen | Cp | 83.862 | 88.141 | 74.433 |
| | Cp + AEM | 85.746 | 88.780 | 76.268 |
| | Cp + Sp + AEM(Sum) | 86.575 | 89.831 | 77.529 |
| | Cp + Sp + AEM(Cat) | 87.059 | 89.715 | 78.779 |
| | Cp + Sp + AEM + FAM | 89.497 | 90.681 | 81.833 |
| Potsdam | Cp | 89.716 | 87.912 | 84.354 |
| | Cp + AEM | 90.600 | 89.275 | 85.864 |
| | Cp + Sp + AEM(Sum) | 91.029 | 89.368 | 86.450 |
| | Cp + Sp + AEM(Cat) | 91.233 | 89.819 | 86.912 |
| | Cp + Sp + AEM + FAM | 92.498 | 91.095 | 88.561 |

specially design an attention enhancement module (AEM) in the contextual path. As presented in Table I, for two datasets, the utilization of AEM (indicated as *Cp + AEM*) brings more than 1.5% improvement in mIoU.

*Ablation for the spatial path*: As the affluent spatial information is crucial for semantic segmentation, the spatial path is designed for preserving the spatial size and extracting spatial information. Table I demonstrated that even the simple fusion schemes such as summation (represented as *Cp + Sp + AEM(Sum)*) and concatenation (represented as *Cp + Sp + AEM(Cat)*) boost the performance.

TABLE II

THE COMPLEXITY AND SPEED OF THE PROPOSED ABCNET AND COMPARATIVE METHODS.

| Method | Backbone | Complexity(G) | Parameters(M) | 256×256 | 512×512 | 1024×1024 | 2048×2048 | 4096×4096 | mIoU |
|---|---|---|---|---|---|---|---|---|---|
| DABNet (Li et al., 2019) | - | 5.22 | 0.75 | 90.67 | 87.74 | 27.41 | 7.44 | * | 82.144 |
| ERFNet (Romera et al., 2017) | - | 14.75 | 2.06 | 90.51 | 59.04 | 17.59 | 4.87 | 1.25 | 79.152 |
| BiSeNetV1 (Yu et al., 2018) | ResNet18 | 15.25 | 13.61 | 143.50 | 87.63 | 25.89 | 7.23 | 1.84 | 84.537 |
| PSPNet (Zhao et al., 2017) | ResNet18 | 12.55 | 24.03 | 151.12 | 105.03 | 34.83 | 10.16 | 2.66 | 77.971 |
| BiSeNetV2 (Yu et al., 2020) | - | 13.91 | 12.30 | 124.49 | 82.84 | 25.64 | 7.07 | * | 85.167 |
| DANet (Fu et al., 2019) | ResNet18 | 9.90 | 12.68 | 181.66 | 124.18 | 40.80 | 11.42 | * | 82.546 |
| FANet (Hu et al., 2020) | ResNet18 | 21.66 | 13.81 | 112.59 | 67.97 | 20.41 | 5.57 | * | 86.722 |
| ShelfNet (Zhuang et al., 2019) | ResNet18 | 12.36 | 14.58 | 123.59 | 90.41 | 30.93 | 9.06 | 2.40 | 86.770 |
| SwiftNet (Oršić and Šegvić, 2021) | ResNet18 | 13.08 | 11.80 | 157.63 | 97.62 | 30.79 | 8.65 | * | 86.285 |
| MAResU-Net (Li et al., 2020a) | ResNet18 | 25.43 | 16.17 | 70.12 | 37.55 | 13.35 | 3.51 | * | 85.928 |
| EaNet (Zheng et al., 2020) | ResNet18 | 18.75 | 34.23 | 73.98 | 55.95 | 17.94 | 5.53 | 1.54 | 85.763 |
| ABCNet | ResNet18 | 18.72 | 14.06 | 113.09 | 72.13 | 22.73 | 6.23 | 1.60 | 88.561 |

* means the network is out of memory.

*Ablation for feature aggregation module*: Given the features obtained by the spatial path and the contextual path are in different domains, neither summation nor the concatenation is the optimal fusion scheme. As can be seen from Table I, the significant gap of performance explains the validity of the feature aggregation module (signified as *Cp + Sp + AEM + FAM*).

## 5) The complexity and speed of the network

The complexity and speed are momentous factors for measuring the merit of an algorithm, which is especially true for practical application. For a thorough comparison, we implement our experiments under different settings. First, the comparison of parameters and computational complexity between different networks are reported in Table II, where 'G' indicates Gillion (i.e., the unit of floating point operations) and 'M' signifies Million (i.e., the unit of parameter number). Meanwhile, for a fair comparison, we choose 256×256, 512×512, 1024×1024, 2048×2048, and 4096×4096 as resolutions of the input image and report the inference speed which is measured by frames per second (FPS) on a midrange notebook graphics card 1660Ti.

The proposed ABCNet simultaneously juggles both speed and accuracy. As can be seen from the last column of Table II, the mIoU on the Potsdam dataset achieved by the ABCNet is at least 1.79% higher than the comparative methods. Meanwhile, the ABCNet could maintain a 72.13 FPS speed for a 512×512 input. Besides, the elaborate design enables the ABCNet to handle the massive input (4096×4096), while more than half of the comparative methods run out of memory for a such large input.

## 6) Results on the ISPRS Vaihingen dataset

The ISPRS Vaihingen is a relatively small dataset. Besides, there is a small covariate shift between training and test sets (Ghassemi et al., 2019). Therefore, the high performance can be easily achieved by specifically designed networks, especially for those fuse orthophoto (TOP) images with auxiliary DSM or NDSM. In this part, we will show that our ABCNet model using

TABLE III

QUANTITATIVE COMPARISON RESULTS ON THE VAIHINGEN TEST SET.

| Method | Backbone | Imp. surf. | Building | Low veg. | Tree | Car | Mean F1 | OA (%) | mIoU (%) |
|---|---|---|---|---|---|---|---|---|---|
| DABNet (Li et al., 2019) | - | 87.775 | 88.808 | 74.319 | 84.905 | 60.247 | 79.211 | 84.278 | 67.373 |
| ERFNet (Romera et al., 2017) | - | 88.451 | 90.239 | 76.394 | 85.751 | 53.649 | 78.897 | 85.751 | 67.698 |
| BiSeNetV1 (Yu et al., 2018) | ResNet18 | 89.115 | 91.304 | 80.867 | 86.911 | 73.122 | 84.264 | 87.084 | 74.094 |
| PSPNet (Zhao et al., 2017) | ResNet18 | 89.005 | 93.161 | 81.483 | 87.657 | 43.926 | 79.046 | 87.651 | 68.861 |
| BiSeNetV2 (Yu et al., 2020) | - | 89.884 | 91.911 | 82.020 | 88.271 | 71.417 | 84.701 | 87.972 | 75.005 |
| DANet (Fu et al., 2019) | ResNet18 | 89.983 | 93.879 | 82.218 | 87.301 | 44.540 | 79.584 | 88.150 | 69.596 |
| FANet (Hu et al., 2020) | ResNet18 | 90.652 | 93.782 | 82.595 | 88.555 | 71.602 | 85.437 | 88.872 | 75.884 |
| EaNet (Zheng et al., 2020) | ResNet18 | 91.675 | 94.522 | 83.095 | 89.243 | 79.984 | 87.704 | 89.688 | 79.223 |
| ShelfNet (Zhuang et al., 2019) | ResNet18 | 91.825 | 94.562 | 83.776 | 89.270 | 77.906 | 87.468 | 89.806 | 78.943 |
| MAResU-Net (Li et al., 2020a) | ResNet18 | 91.971 | 95.044 | 83.735 | 89.349 | 78.283 | 87.676 | 90.047 | 80.749 |
| SwiftNet (Oršić and Šegvić, 2021) | ResNet18 | 92.222 | 94.843 | 84.138 | 89.309 | 81.234 | 88.349 | 90.199 | 80.034 |
| ABCNet | ResNet18 | **92.726** | **95.239** | **84.541** | **89.680** | **85.299** | **89.497** | **90.681** | **81.833** |

only TOP images with efficient architecture can not only also transcend lightweight networks but also achieve competitive performance with those specially designed models.

As shown in TABLE III, the numeric scores for the ISPRS Vaihingen test dataset demonstrated that our ABCNet delivers robust performance, and exceeded other lightweight networks in the mean F1, OA, and mIoU by a considerable margin. Significantly, the "car" class in Vaihingen dataset is difficult to handle as it is a relatively small object. Nonetheless, our ABCNet acquires

TABLE IV

KAPPA Z-TEST COMPARING THE PERFORMANCE OF DIFFERENT METHODS ON THE VAIHINGEN DATASET.

| Method | Kappa | KV | ERFNet | PSPNet | BiSeNetV1 | DANet | BiSeNetV2 | FANet | EaNet | ShelfNet | MAResU-Net | SwiftNet | ABCNet |
|---|---|---|---|---|---|---|---|---|---|---|---|---|---|
| DABNet | 0.798 | 2.808 | 6.04 | 19.84 | 21.73 | 22.53 | 26.86 | 27.89 | 33.03 | 34.38 | 35.68 | 35.94 | 39.06 |
| ERFNet | 0.812 | 2.643 | - | 13.80 | 15.70 | 16.50 | 20.84 | 21.86 | 27.02 | 28.37 | 29.67 | 29.93 | 33.06 |
| BiSeNetV1 | 0.843 | 2.272 | - | - | 1.89 | 2.70 | 7.04 | 8.08 | 13.24 | 14.60 | 15.91 | 16.17 | 19.32 |
| PSPNet | 0.847 | 2.218 | - | - | - | 0.80 | 5.15 | 6.19 | 11.35 | 12.72 | 14.03 | 14.29 | 17.44 |
| BiSeNetV2 | 0.849 | 2.198 | - | - | - | - | 4.35 | 5.38 | 10.55 | 11.91 | 13.22 | 13.48 | 16.63 |
| DANet | 0.858 | 2.081 | - | - | - | - | - | 1.04 | 6.21 | 7.57 | 8.88 | 9.14 | 12.30 |
| FANet | 0.860 | 2.057 | - | - | - | - | - | - | 5.17 | 6.53 | 7.84 | 8.10 | 11.26 |
| EaNet | 0.870 | 1.918 | - | - | - | - | - | - | - | 1.36 | 2.68 | 2.94 | 6.10 |
| ShelfNet | 0.873 | 1.883 | - | - | - | - | - | - | - | - | 1.31 | 1.57 | 4.73 |
| MAResU-Net | 0.875 | 1.850 | - | - | - | - | - | - | - | - | - | 0.26 | 3.42 |
| SwiftNet | 0.876 | 1.843 | - | - | - | - | - | - | - | - | - | - | 3.16 |
| ABCNet | 0.882 | 1.762 | - | - | - | - | - | - | - | - | - | - | - |

an 85.299% F1 score, which is at least 4% higher than other methods. To further evaluate the statistical significance, we report Kappa z-test for pairwise methods based on Kappa coefficients of agreement and their variances using the following equation:

$$z = (k_1 - k_1)/\sqrt{v_1 + v_2}, \tag{20}$$

where $k$ signifies the Kappa coefficient and $v$ denotes the Kappa variance. Concretely, if the value of $z$ is greater than 1.96, the two algorithms are signally different at the 95 % confidence level.

TABLE V

QUANTITATIVE COMPARISON RESULTS ON THE VAIHINGEN TEST SET WITH STATE-OF-THE-ART METHODS.

| Method | Backbone | Imp. surf. | Building | Low veg. | Tree | Car | Mean F1 | OA (%) | mIoU (%) | Speed |
|---|---|---|---|---|---|---|---|---|---|---|
| DeepLabV3+ (Chen et al., 2018a) | ResNet101 | 92.38 | 95.17 | 84.29 | 89.52 | 86.47 | 89.57 | 90.56 | 81.47 | 13.27 |
| PSPNet (Zhao et al., 2017) | ResNet101 | 92.79 | 95.46 | 84.51 | 89.94 | **88.61** | 90.26 | 90.85 | **82.58** | 22.03 |
| DANet (Fu et al., 2019) | ResNet101 | 91.63 | 95.02 | 83.25 | 88.87 | 87.16 | 89.19 | 90.44 | 81.32 | 21.97 |
| EaNet (Zheng et al., 2020) | ResNet101 | **93.40** | **96.20** | 85.60 | 90.50 | 88.30 | **90.80** | 91.20 | - | 9.97 |
| DDCM-Net (Liu et al., 2020) | ResNet50 | 92.70 | 95.30 | 83.30 | 89.40 | 88.30 | 89.80 | 90.40 | - | 37.28 |
| HUSTW5 (Sun et al., 2019) | ResegNets | 93.30 | 96.10 | **86.40** | **90.80** | 74.60 | 88.20 | **91.60** | - | - |
| CASIA2 (Liu et al., 2018) | ResNet101 | 93.20 | 96.00 | 84.70 | 89.90 | 86.70 | 90.10 | 91.10 | - | - |
| V-FuseNet# (Audebert et al., 2018) | FuseNet | 91.00 | 94.40 | 84.50 | 89.90 | 86.30 | 89.20 | 90.00 | - | - |
| DLR_9# (Marmanis et al., 2018) | - | 92.40 | 95.20 | 83.90 | 89.90 | 81.20 | 88.50 | 90.30 | - | - |
| ABCNet | ResNet18 | 92.73 | 95.24 | 84.54 | 89.68 | 85.30 | 89.50 | 90.68 | 81.83 | **72.13** |

- means the results are not repoted in the original paper.

# means the DSM or NDSM are used in the network.

As can be seen from Table IV, the accuracy of the proposedABCNet is statistically higher than other comparative methods. In addition, we visualize area 38 in Fig. 5 to qualitatively demonstrate the effectiveness of our ABCNet, while the enlarged results are shown in Fig. 7 (a).

For a comprehensive evaluation, ABCNet is also compared with other state-of-the-art methods. As can be seen in Table V, as a lightweight network, the proposed ABCNet achieves a competitive performance even compared with those designed models with complex structures. It is worth noting that the speed of our ABCNet is two to seven times faster than those methods.

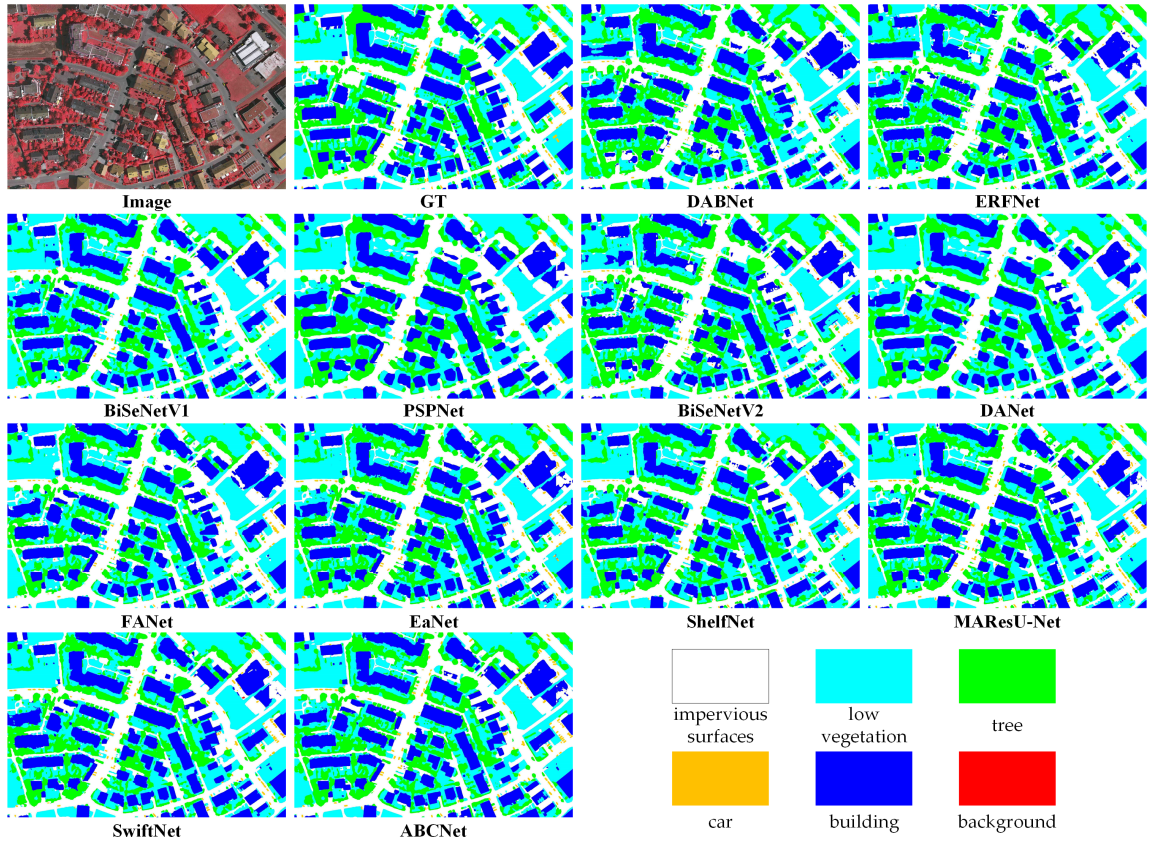

Fig.5 Mapping results for test images of Vaihingen tile-38.

## 7) Results on the ISPRS Potsdam dataset

We carry out experiments on the ISPRS Potsdam dataset to further evaluate the performance of ABCNet. Numerical comparisons with other lightweight methods are shown in Table VI, while the Kappa-z test is illustrated in Table VII. Remarkably, ABCNet achieves 91.095% in overall accuracy and 88.561% in mIoU, and the Kappa-z test strongly illuminates the superiority contrasted with other lightweight networks. The visualization of area 3_13 is displayed in Fig. 6, and the enlarged results are exhibited in Fig. 7 (b). As there are sufficient images in the Potsdam dataset to train the network, the performance of the ABCNet can be parity with the state-of-the-

TABLE VI

QUANTITATIVE COMPARISON RESULTS ON THE POTSDAM TEST SET.

| Method | Backbone | Imp. surf. | Building | Low veg. | Tree | Car | Mean F1 | OA (%) | mIoU (%) |
|---|---|---|---|---|---|---|---|---|---|
| ERFNet (Romera et al., 2017) | - | 88.675 | 92.991 | 81.100 | 75.843 | 90.534 | 85.829 | 84.492 | 79.152 |
| DABNet (Li et al., 2019) | - | 89.939 | 93.188 | 83.596 | 82.257 | 92.578 | 88.312 | 86.664 | 82.144 |
| PSPNet (Zhao et al., 2017) | ResNet18 | 89.116 | 94.501 | 84.041 | 85.766 | 76.622 | 86.009 | 87.216 | 77.971 |
| BiSeNetV1 (Yu et al., 2018) | ResNet18 | 90.241 | 94.554 | 85.527 | 86.195 | 92.684 | 89.840 | 88.163 | 84.537 |
| BiSeNetV2 (Yu et al., 2020) | - | 91.280 | 94.316 | 85.048 | 85.192 | 94.112 | 89.990 | 88.174 | 85.167 |
| EaNet (Zheng et al., 2020) | ResNet18 | 92.008 | 95.692 | 84.308 | 85.719 | 95.112 | 90.568 | 88.703 | 85.763 |
| MAResU-Net (Li et al., 2020a) | ResNet18 | 91.414 | 95.572 | 85.823 | 86.608 | 93.306 | 90.545 | 89.043 | 85.928 |
| DANet (Fu et al., 2019) | ResNet18 | 91.003 | 95.567 | 86.089 | 87.579 | 84.301 | 88.908 | 89.129 | 82.546 |
| SwiftNet (Oršić and Šegvić, 2021) | ResNet18 | 91.834 | 95.943 | 85.721 | 86.837 | 94.456 | 90.958 | 89.329 | 86.285 |
| FANet (Hu et al., 2020) | ResNet18 | 91.985 | 96.101 | 86.045 | 87.833 | 94.533 | 91.299 | 89.822 | 86.722 |
| ShelfNet (Zhuang et al., 2019) | ResNet18 | 92.530 | 95.750 | 86.595 | 87.070 | 94.585 | 91.306 | 89.920 | 86.770 |
| ABCNet | ResNet18 | **93.270** | **96.798** | **87.814** | **88.687** | **95.921** | **92.498** | **91.095** | **88.561** |

art methods with a much faster speed. The comparisons are illustrated in Table VIII.

# 5.  CONCLUSIONS

In this paper, we propose a novel lightweight framework for efficient semantic segmentation

in the field of remote sensing, namely Attentive Bilateral Contextual Network (ABCNet), which

adaptively captures abundant spatial details by spatial path and global contextual information via

TABLE VII

KAPPA Z-TEST COMPARING THE PERFORMANCE OF DIFFERENT METHODS ON THE POTSDAM DATASET.

| Method | Kappa | KV | DABNet | PSPNet | BiSeNetV1 | BiSeNetV2 | EaNet | DANet | MAResU-Net | SwiftNet | FANet | ShelfNet | ABCNet |
|--------|-------|-----|--------|--------|-----------|-----------|-------|-------|------------|----------|-------|----------|--------|
| ERFNet | 0.837 | 4.344 | 9.06 | 11.25 | 17.17 | 17.51 | 19.64 | 20.18 | 21.01 | 22.27 | 23.84 | 24.32 | 29.66 |
| DABNet | 0.863 | 3.712 | - | 2.19 | 8.14 | 8.50 | 10.64 | 11.19 | 12.02 | 13.29 | 14.88 | 15.37 | 20.77 |
| PSPNet | 0.869 | 3.563 | - | - | 5.96 | 6.33 | 8.46 | 9.01 | 9.85 | 11.12 | 12.71 | 13.21 | 18.62 |
| BiSeNetV1 | 0.884 | 3.187 | - | - | - | 0.37 | 2.50 | 3.06 | 3.90 | 5.16 | 6.76 | 7.26 | 12.69 |
| BiSeNetV2 | 0.885 | 3.182 | - | - | - | - | 2.13 | 2.68 | 3.52 | 4.78 | 6.38 | 6.88 | 12.30 |
| EaNet [7] | 0.890 | 3.032 | - | - | - | - | - | 0.56 | 1.40 | 2.66 | 4.26 | 4.77 | 10.20 |
| DANet | 0.892 | 3.006 | - | - | - | - | - | - | 0.84 | 2.10 | 3.70 | 4.21 | 9.64 |
| MAResU-Net | 0.894 | 2.959 | - | - | - | - | - | - | - | 1.26 | 2.86 | 3.37 | 8.80 |
| SwiftNet | 0.897 | 2.870 | - | - | - | - | - | - | - | - | 1.60 | 2.11 | 7.54 |
| FANet | 0.901 | 2.780 | - | - | - | - | - | - | - | - | - | 0.51 | 5.94 |
| ShelfNet | 0.902 | 2.757 | - | - | - | - | - | - | - | - | - | - | 5.43 |
| ABCNet | 0.914 | 2.425 | - | - | - | - | - | - | - | - | - | - | - |

the contextual path. In particular, we design an attention enhancement module to model long-range dependencies from extracted feature maps. Additionally, to address the feature fusion issue and improve the effectiveness, a feature aggregation module is presented to adequately merge the detailed features captured by the spatial path and semantic features generated by the contextual path. Extensive experiments on ISPRS Vaihingen and Potsdam datasets demonstrate the effectiveness and efficiency of the proposed ABCNet.

TABLE VIII

QUANTITATIVE COMPARISON RESULTS ON THE POTSDAM TEST SET WITH STATE-OF-THE-ART METHODS.

| Method | Backbone | Imp. surf. | Building | Low veg. | Tree | Car | Mean F1 | OA (%) | mIoU (%) | Speed |
|---|---|---|---|---|---|---|---|---|---|---|
| DeepLabV3+ (Chen et al., 2018a) | ResNet101 | 92.95 | 95.88 | 87.62 | 88.15 | 96.02 | 92.12 | 90.88 | 84.32 | 13.27 |
| PSPNet (Zhao et al., 2017) | ResNet101 | 93.36 | 96.97 | 87.75 | 88.50 | 95.42 | 94.40 | 91.08 | 84.88 | 22.03 |
| DDCM-Net (Liu et al., 2020) | ResNet50 | 92.90 | 96.90 | 87.70 | 89.40 | 94.90 | 92.30 | 90.80 | - | 37.28 |
| CCNet (Huang et al., 2020) | ResNet101 | 93.58 | 96.77 | 86.87 | 88.59 | 96.24 | 92.41 | 91.47 | 85.65 | 5.56 |
| AMA_1 | - | 93.40 | 96.80 | 87.70 | 88.80 | 96.00 | 92.54 | 91.20 | - | - |
| SWJ_2 | ResNet101 | 94.40 | 97.40 | 87.80 | 87.60 | 94.70 | 92.38 | 91.70 | - | - |
| HUSTW4 (Sun et al., 2019) | ResegNets | 93.60 | 97.60 | 88.50 | 88.80 | 94.60 | 92.62 | 91.60 | - | - |
| V-FuseNet# (Audebert et al., 2018) | FuseNet | 92.70 | 96.30 | 87.30 | 88.50 | 95.40 | 92.04 | 90.60 | | |
| DST_5# (Sherrah, 2016) | FCN | 92.50 | 96.40 | 86.70 | 88.00 | 94.70 | 91.66 | 90.30 | | |
| ABCNet | ResNet18 | 93.27 | 96.80 | 87.81 | 88.69 | 95.92 | 92.50 | 91.10 | 88.56 | 72.13 |

- means the results are not repoted in the original paper.

\# means the DSM or NDSM are used in the network.

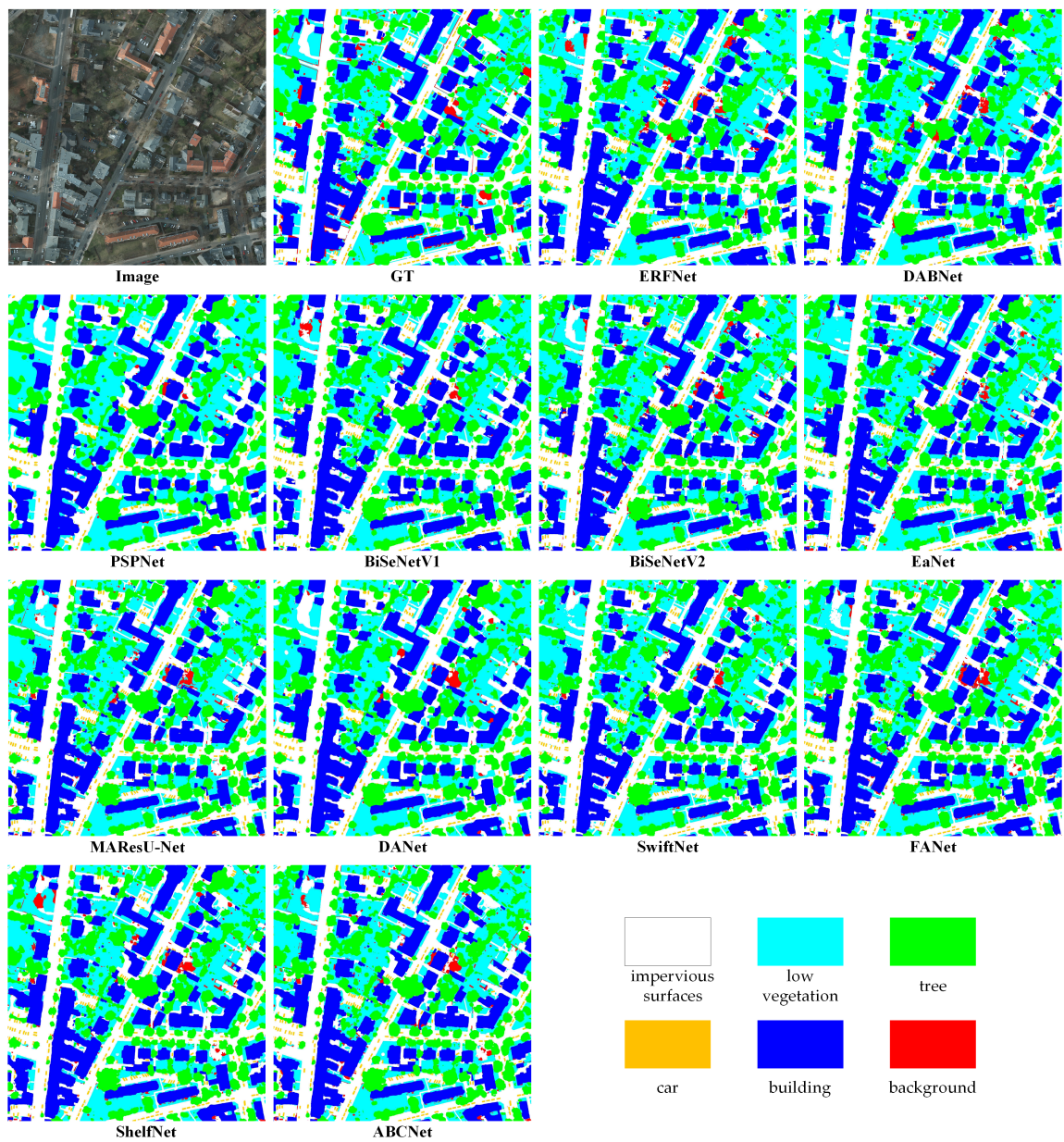

Fig.6 Mapping results for test images of Potsdam tile-3_13.

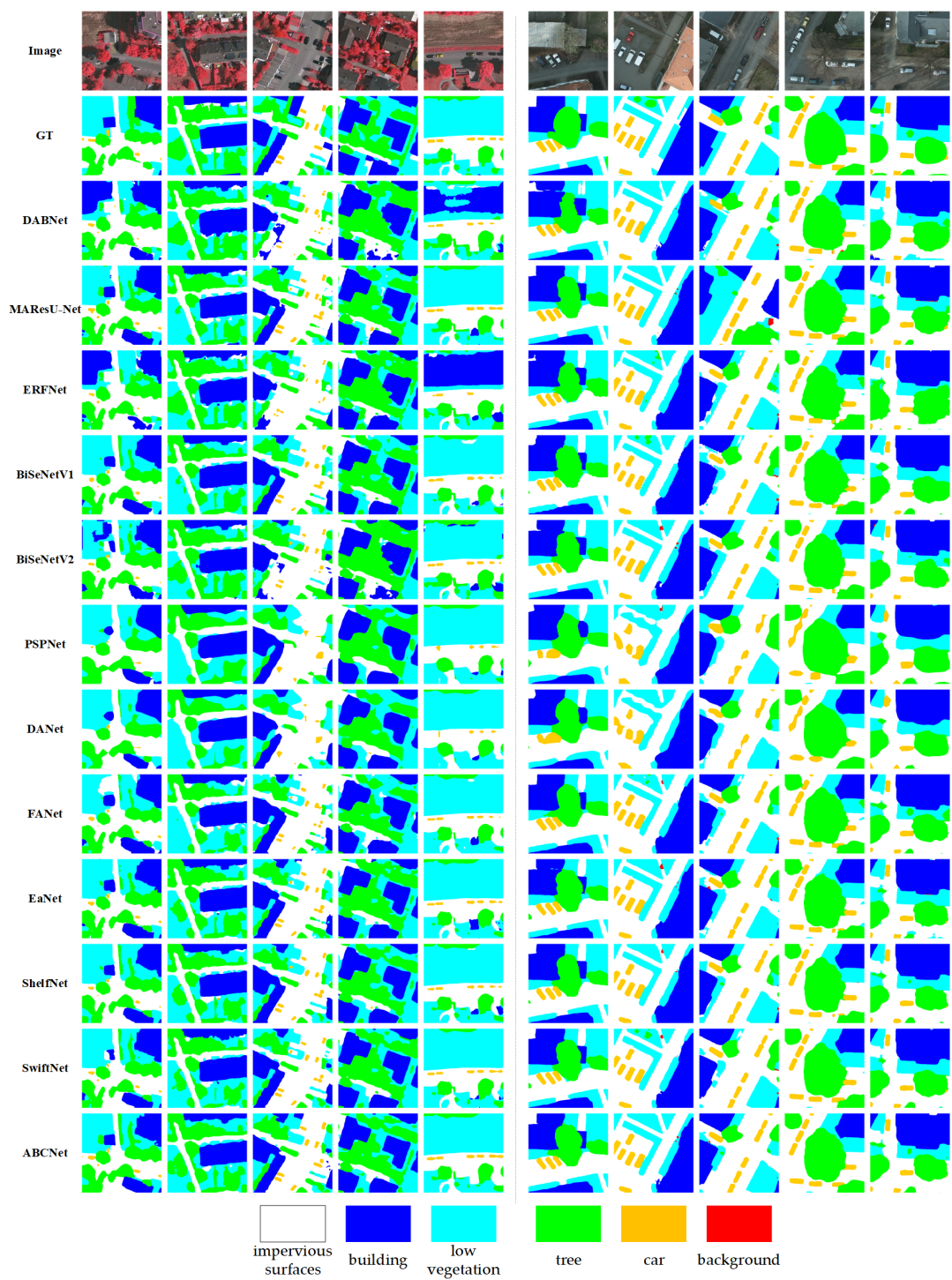

Fig.7 Enlarged visualization of results on (LEFT) the Vaihingen dataset and (RIGHT)

Potsdam dataset.